\documentclass{article}

\PassOptionsToPackage{numbers,sort&compress}{natbib}
\usepackage{float}
\usepackage{graphicx}
\usepackage[preprint]{neurips_2026}
\usepackage[utf8]{inputenc} 
\usepackage[T1]{fontenc}    
\usepackage{hyperref}       
\usepackage{url}            
\usepackage{booktabs}       
\usepackage{amsfonts}       
\usepackage{nicefrac}       
\usepackage{microtype}      
\usepackage{xcolor}         

\usepackage{makecell}

\usepackage{amsmath}

\title{Geometry-aware Latent Autoregressive Generative Model for PDEs in Complex Domains}

\author{
    Zi Wang\thanks{Equal contribution.}\thanks{Corresponding author.} \quad Minghui Xu\footnotemark[1] \quad Tapan Mukerji \\
    Department of Energy Science and Engineering \\
    Stanford University \\
    Stanford, CA 94305 \\
    \texttt{\{ziwang3, minghuix, mukerji\}@stanford.edu} \\
  }

\begin{document}

\renewcommand{\thefootnote}{\fnsymbol{footnote}}
\maketitle
\renewcommand{\thefootnote}{\arabic{footnote}}

\begin{abstract}
  Solving multiphysics partial differential equations (PDEs) remains a major challenge in scientific computing, especially for highly complex $\mu$m-scale tortuous geometries critical to energy and chemical engineering. We address this challenge by proposing a Geometry-aware Latent Autoregressive generative Model for PDEs (GeoLAMP), which solves physics within highly irregular and tortuous structures by decoupling flow and transport physics. GeoLAMP introduces a dual-encoder architecture on graph representations to jointly capture global topology and fine-scale geometric features, enabling an effective transition from real-space fields to compact latent representations. In the latent space, we propose a causal self-attention transformer with flow matching to model temporal dynamics, allowing stable and scalable block-wise autoregressive prediction. In addition, we propose a grid–graph data fusion scheme that projects low-resolution grid-based approximate priors onto graph representations, improving prediction of flow in tortuous structures. We establish three multiphysics benchmark datasets in complex geometries, covering reactive flow, heat convection, and elasticity. GeoLAMP consistently achieves the most stable autoregression performance on these datasets. Our results provide a systematic study of geometry-aware learning for PDEs in $\mu$m-scale complex geometries and offer new insights into block-wise time marching of latent autoregressive PDE modeling via a flow matching framework.
\end{abstract}

\section{Introduction}
Scientific machine learning for partial differential equations (PDEs) has gained substantial attention and has multiple methodological directions, ranging from physics-informed fitting  \citep{raissi2019physics,karniadakis2021physics,pang2019fpinns} to operator learning  \citep{li2021fno,lu2021deeponet,li2023oformer,azizzadenesheli2024neural} and latent modeling  \citep{wu2022learning,iakovlev2023learning,kontolati2024latent}. It shows strong potential to reshape conventional computational workflows in many scientific and engineering scenarios, including weather forecasting  \citep{chen2024machine}, turbulence reconstruction  \citep{wang2020towards}, and geological engineering  \citep{wen2022u}. However, compared with the widely studied physical problems, multiphysics processes occurring inside complex meso-scale structures ($\mu$m) have received far less attention  \citep{chen2022pore}. More importantly, meso-scale physics often governs macroscopic scientific and engineering behavior, such as natural hydrogen  \citep{mathur2025changes}, chip cooling \citep{van2020co} and battery design  \citep{lu20203d}. This cross-scale dependence poses significant computational challenges. Therefore, it is necessary to develop and apply deep learning methods specifically tailored to physical systems defined on meso-scale complex geometries  \citep{li2013multiscale,agrawal2001role}.

\begin{figure}[t]
    \centering
    \includegraphics[width=0.95\linewidth]{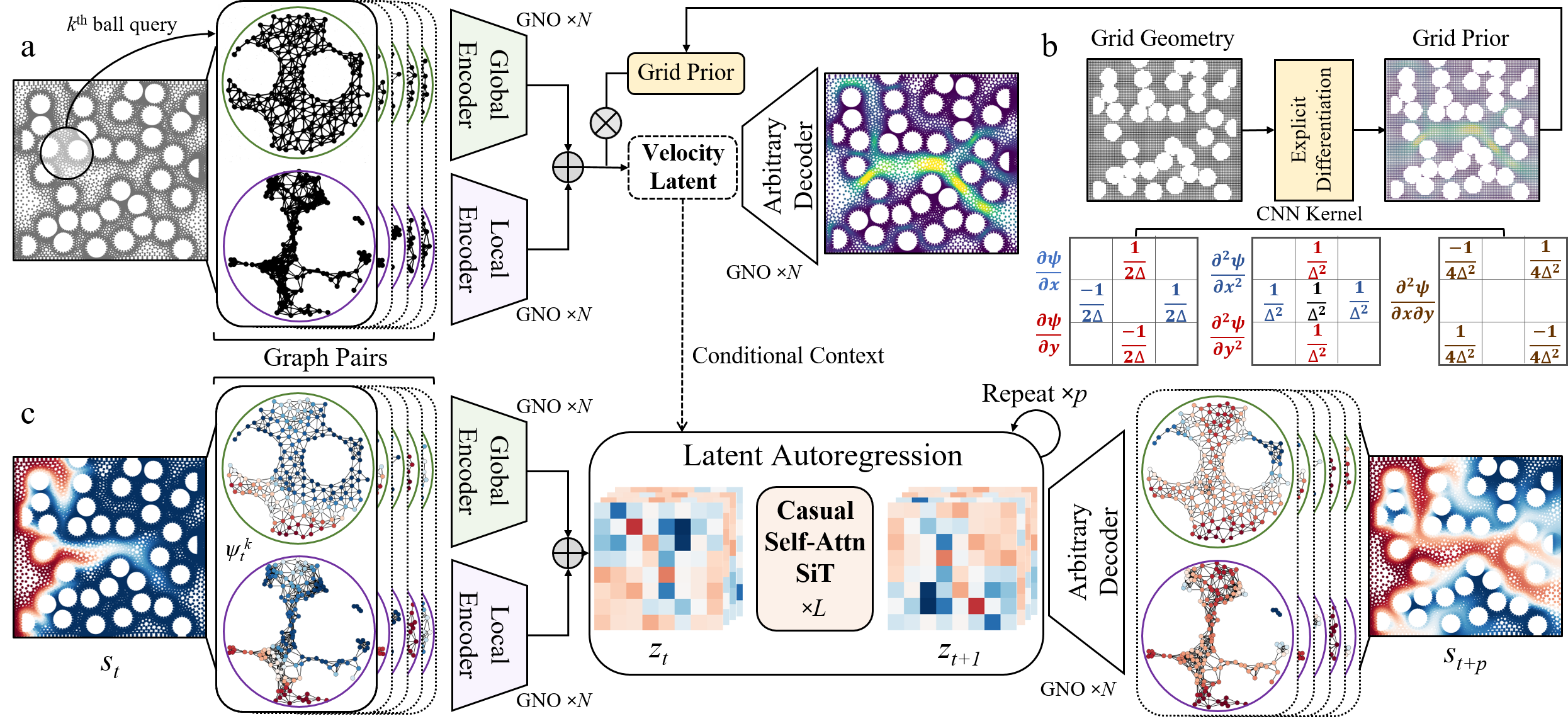}
    \caption{Schematic of GeoLAMP architecture. (a) Prediction of tortuous fluid flow on graph pairs. (b) Grid-based flow prior generation. (c) Autoregression of time-evolved physics.}
    \label{geolamp_schm}
    \vspace{-10pt}
\end{figure}
To handle complex geometries, several machine-learning-based studies have proposed efficient solutions by incorporating geometric information into model architectures  \citep{bronstein2021geometric,brandstetter2022mppde,takamoto2022pdebench}. Point-cloud methods represent complex geometries as unordered sets of spatial points, thereby avoiding the need for structured grids  \citep{qi2017pointnetpp,li2023oformer,wu2024transolver,kashefi2021point,liang2013solving,wang2019dgcnn}. Graph-based methods further augment these representations with explicit neighborhood connectivity, enabling message passing or graph-kernel operators to model local physical interactions and long-range geometry-dependent transport  \citep{li2020gkn,li2023gino,pfaff2021meshgraphnets,sanchezgonzalez2020simulate,chung2024prediction}. More recently, GeoPT  \citep{wu2026geopt} introduces lifted geometric pre-training, which augments large-scale geometry data with synthetic dynamics to bridge the gap between static shape representation and physics simulation. However, real-space geometric complexity increases dramatically with domain size and problem scale, especially for meso-scale structures represented by millions of voxels.

To address this difficulty, latent-space methods  \citep{gao2026gencp,li2025latent,wu2022learning,kontolati2024latent} encode real-space data into compact latent representations and model temporal evolution in latent space, leading to substantially improved efficiency and flexibility through compression. For complex domains, latent modeling is more appealing if the geometry-aware backbones can compress irregular point clouds or meshes into regular and compact latent representations while preserving real-space information  \citep{wang2019dgcnn,li2020gkn}. Meso-scale geometries characterized by highly tortuous and heterogeneous structures calls for more reffned geometry-aware encoder–decoder designs that can compress complex domains while preserving geometry-sensitive physical information.

Once the dynamics are represented in a latent space, diffusion-based generative modeling    \citep{ho2020ddpm,lipman2023flowmatching,peebles2023dit,liu2023rectifiedflow,nichol2021improvedddpm,song2021score,song2021ddim,karras2022edm} provides a principled paradigm for accurate and stable temporal prediction. At scale, latent tokenization and transformer backbones have further shown that high-dimensional generation becomes more efficient when the model operates in a compact latent space  \citep{oord2017vqvae,esser2021taming,rombach2022ldm,peebles2023dit,ma2024sit}. These ideas are increasingly relevant to scientific surrogate modeling \citep{lippe2023pderefiner,hao2024dpot,li2025latent}.

In this study, we propose a \textbf{Geo}metry-aware \textbf{L}atent \textbf{A}utoregressive generative \textbf{M}odel for \textbf{P}DEs (GeoLAMP) to learn time-dependent multiphysics behavior in complex meso-scale geometries through latent representations, as shown in Fig.~\ref{geolamp_schm}. Specifically, GeoLAMP consists of three main components: $(i)$ Global and Local Encoders (GE \& LE) uses multiple graph neural operator (GNO) layers to encode global/local information from graph pairs of query points sampled by the farthest-point sampling (FPS) and covariance-based sampling (CBS), separately; $(ii)$ Causal Self-attention Transformer performs block-wise temporal autoregression by applying masked self-attention over timestep tokens; and $(iii)$ A grid–graph fusion scheme embeds a convolution-derived tortuous flow velocity prior into the graph structure, enhancing prediction by embedding physical priors. GeoLAMP first encodes real-space information into a latent space using GE and LE. The physical velocity field is then predicted from pure geometry and prior information, and the temporal dynamics are autoregressively generated from the initial latent state conditioned on it. Finally, the generated latent representations are decoded back to real space at arbitrary spatial locations and time points. We constructed three meso-scale physics datasets. GeoLAMP achieves the most stable autoregression behavior on them, maintaining low errors throughout the entire rollout horizon.

\vspace{-8pt}
\paragraph{Contributions} (i) A dual-GNO encoder combines farthest-point coverage with covariance-selected interior detail on irregular meshes, while a coarse-grid flow prior supports geometry-to-velocity prediction. (ii) A target-isolated causal mask lets flow matching generate an entire latent block in parallel, with autoregression across committed blocks. Shared frame-wise parameters allow a frozen model to vary both history length $C$ and one-call output length $P$ by rebuilding the mask; Section~\ref{Inferencetimerolloutflexibility} quantifies the accuracy and serial-call tradeoff. (iii) We evaluate the design on reactive flow, heat convection, and elasticity in complex geometries, including geometry baselines and a matched direct-regression control. 

\section{Related Work}
Existing datasets for learning PDE dynamics fall short of the regime we
target on at least one axis as shown in Table~\ref{tab:dataset_comparison}. Our three datasets occupy exactly this intersection of tortuous volumetric domains, time-dependent rollout and multiphysics coupling, and this regime motivates two design choices that distinguish GeoLAMP from prior work.
\begin{table}[H]
\centering
\small
\caption{Comparison of widely used AI-for-PDE datasets with ours.}
\label{tab:dataset_comparison}
\begin{tabular}{lccc}
\toprule
Dataset & \makecell{Complex geometry} & \makecell{Time-dependent} & \makecell{Multi-physics} \\
\midrule
FNO suite - NS \citep{li2021fno}          & $\times$ & $\checkmark$ & $\times$ \\
PDEBench \citep{takamoto2022pdebench}     & $\times$ & $\checkmark$ & $\times$ \\
The Well \citep{ohana2024well}            & $\times$ & $\checkmark$ & $\checkmark$ \\
Geo-FNO Elasticity \citep{li2023geofno}   & $\checkmark$ (Single object) & $\times$ & $\times$ \\
Geo-FNO Airfoil/Pipe \citep{li2023geofno} & $\checkmark$ (Single object) & $\times$ & $\times$ \\
ShapeNet-Car \citep{umetani2018car}       & $\checkmark$ (Single object) & $\times$ & $\times$ \\
AirfRANS \citep{bonnet2022airfrans}       & $\checkmark$ (Single object) & $\times$ & $\times$ \\
Ahmed-Body \citep{li2023gino}             & $\checkmark$ (Single object) & $\times$ & $\times$ \\
\midrule
\textbf{Ours (3 datasets)} & $\checkmark$ (Tortuous domain)& $\checkmark$ & $\checkmark$ \\
\bottomrule
\end{tabular}
\end{table}

The first concerns geometry representation. Existing geometry-aware solvers
convert an irregular shape into a grid-compatible form: Geo-FNO
\citep{li2023geofno} learns a diffeomorphic deformation to a latent regular
grid, NUNO \citep{liu2023nuno} partitions the domain into individually
gridded subdomains, GINO \citep{li2023gino} projects the mesh onto a regular
latent grid through GNO layers, and attention-based solvers
\citep{li2023oformer, wu2024transolver} operate directly on mesh
points or learned slices. These designs target domains dominated by a single
smooth boundary such as an airfoil, a car body and a pipe, whereas tortuous
meso-scale domains are multiply connected with tens of internal boundaries. GeoLAMP therefore
represents the whole volumetric domain rather than a surface: global encoder provides globally uniform coverage, and local encoder densifies pore and channel interiors. Together, the dual encoder effectively recover the physics in tortuous domains.

The second concerns temporal prediction. Building on latent diffusion and
flow matching \citep{rombach2022ldm, peebles2023dit, ma2024sit,
lipman2023flowmatching}, denoising-based rollout improves long-horizon
stability over direct regression for PDEs \citep{lippe2023pderefiner,
hao2024dpot, rozet2025lost}; the latent flow-matching solver of
\citet{li2025latent}, the closest generative relative, performs next-step
generation on a regular latent grid. GeoLAMP-B instead generates a block of
$P$ frames in one flow-matching solve, each noisy target group attending to
all context groups but not to other targets, and shared frame-wise parameters
let a frozen checkpoint change the context and block lengths at inference by
rebuilding the causal mask
(Section~\ref{Inferencetimerolloutflexibility}). Together, the domain-level
encoding and block-wise latent generation address the regime of
Table~\ref{tab:dataset_comparison} that prior benchmarks and solvers do not
cover.

\section{Problem setup in complex geometry}
We focus on solving multiphysics PDEs with $K$ solution variables, denoted by $s^{K}(x,t)$, in a complex domain $\Omega \subseteq \mathbb{R}^{d}$ over time $t\in[0,T]$, subject to boundary conditions $BC(x,t)$ and initial conditions $IC(x)$, as defined in Eq.~\ref{pdegen}. Herein, $\mathcal{L}{a}$ denotes a physics-dependent differential operator parameterized by coefficients or source terms $a$. We construct three datasets covering different physical processes and meso-scale structures: flow-reactive transport in circular-packed porous structures, convective heat transfer in stochastic-field-generated structures, and elasticity in foam structures. All datasets are generated using the finite element method (FEM) in COMSOL \citep{multiphysics1998introduction}, with detailed descriptions provided in Appendix~\ref{Appen_A}.
\begin{equation}
\label{pdegen}
\begin{array}{ll}
\mathcal{L}_{a} \circ s^{K}(x,t)=0, & x \in \Omega\\[3pt]
s^{K}(x,t)=BC(x,t), & x \in \partial\Omega,\; t \in [0,T], \\[3pt]
s^{K}(x,0)=IC(x), & x \in \Omega
\end{array}
\end{equation}

\section{Geometry-aware latent generative model}
\subsection{Encoder and decoder}
\label{gl}
The physical information in real space is organized from unstructured mesh outputs as point sets, denoted by ${x_i, s_i^K}{i=1}^{N}$ at time $t$ on $N$ points. Since the full point set is typically dense and computationally expensive, directly mapping all real-space points to latent representations is inefficient. To address this, we design global and local encoders (GE \& LE) that compress the full field through tailored point selection while preserving as much geometry-sensitive physical information as possible. The arbitrary decoder maps the generated latent representations back to real-space points at arbitrary resolutions. All of them are built upon graph neural operators (GNOs) \citep{li2020gkn}, which aggregate information from points within a query ball $B$ of radius $r_b$ and map it onto a regular latent grid ${y_j, g_j^D}_{j=1}^{M}$ with size $M$ at time $t$ through Eq.~\ref{gno_1}, following a Green's function approximation.
\begin{equation}
\label{gno_1}
g_j(y^0) \approx \sum_{x\in B} \kappa(x, y^0_j)\, s(x)\, \mu(x)
\end{equation}
Then, followed with multiple GNO blocks, the real space information is compressed to latent representations, where in each layer, the approximation follows
\begin{equation}
\label{gno_2}
g_j(y^l) \approx \sum_{y^{l-1}_j\in B^l} \kappa(y^l, y^{l-1}_j)\, s(y_j^{l-1})\, \mu(y_j^{l-1})
\end{equation}
In order to comprehensively capture the global and local information, GE and LE takes different embedded points from real space, and they are trained together as a variational auto-encoders (VAE). The GE\&LE selections are shown in Appendix~\ref{Appen_Pointselect}.
\vspace{-6pt}
\paragraph{Global encoder} The GE intends to cover the full spatial domain and avoids local clustering, providing a robust geometric representation for global information, including inlet, outlet and boundary nodes. It captures domain-scale trends from inlet to outlet and along the boundary. Given the full input point cloud $\{x_i\}_{i=1}^{N}$, where $x_i\in\mathbb{R}^{d}$, GE automatically selects $n_g$ representative points using the farthest point sampling (FPS) \citep{moenning2003fast} as the input. Starting from an initial point $x_{i_1}$, the sampled set at step $t$ is denoted as $S_t=\{x_{i_1},x_{i_2},\ldots,x_{i_t}\}$, at each iteration, the next point is selected by maximizing its minimum Euclidean distance to the existing sampled set:
\begin{equation}
x_{i_{t+1}}
=
\arg\max_{x\in P\setminus S_t}
\left(
\min_{x_j\in S_t}
\|x-x_j\|_2
\right).
\end{equation}
This greedy procedure approximately solves the max--min sampling objective:
\begin{equation}
\max_{S\subset P,\ |S|=n_g}
\;
\min_{x\in P}
\min_{x_j\in S}
\|x-x_j\|_2 .
\end{equation}

\vspace{-6pt}
\paragraph{Local encoder} From the points not selected by the global encoder, the local encoder (LE) selects $n_l$ additional points to resolve pore and channel interiors. We use covariance-based sampling (CBS): a neighborhood score derived from the point covariance favors locally isotropic point distributions that complement the global FPS set. Given a point set $\{x_i\}_{i=1}^{N}$, where $x_i\in\mathbb{R}^{d}$, for each point $x_i$, its $k$-nearest neighbors are first constructed as
\begin{equation}
\mathcal{N}_k(x_i)
=
\{x_{i_1},x_{i_2},\ldots,x_{i_k}\}.
\end{equation}

The local geometric variation around $x_i$ is characterized by the covariance matrix of neighbor  points:
\begin{equation}
\mathbf{C}_i
=
\frac{1}{k}
\sum_{x_j\in\mathcal{N}_k(x_i)}
(x_j-\bar{x}_i)(x_j-\bar{x}_i)^{\top},
\end{equation}
where the local centroid is defined as
\begin{equation}
\bar{x}_i
=
\frac{1}{k}
\sum_{x_j\in\mathcal{N}_k(x_i)}
x_j.
\end{equation}

By performing eigenvalue decomposition on $\mathbf{C}_i$, we obtain $\lambda_1 \leq \lambda_2 \leq \cdots \leq \lambda_d$, and define a local neighborhood-isotropy score as
\begin{equation}
\kappa_i
=
\frac{\lambda_1}
{\sum_{m=1}^{d}\lambda_m}.
\end{equation}

A larger $\kappa_i$ indicates a more isotropic local neighborhood; it does not directly measure boundary curvature. LE selects the $n_l$ highest-scoring points among those not chosen by FPS. This densifies interior pores while the FPS branch retains domain-wide coverage, as illustrated in Appendix~\ref{Appen_Pointselect}.
\vspace{-6pt}
\paragraph{Arbitrary decoder} The arbitrary decoder (AD) is designed to decode the generated latent representations back to real-space points at arbitrary locations and time steps. To handle meshes with variable numbers of nodes, we introduce a simple masking strategy into the GNO block, where information is propagated only through valid graph nodes while empty nodes are masked out. Moreover, the full field can be reconstructed at any desired resolution through iterative decoding with AD.
\subsection{Flow matching and autoregressive generation}

\begin{figure}[H]
    \centering
    \includegraphics[width=0.85\linewidth]{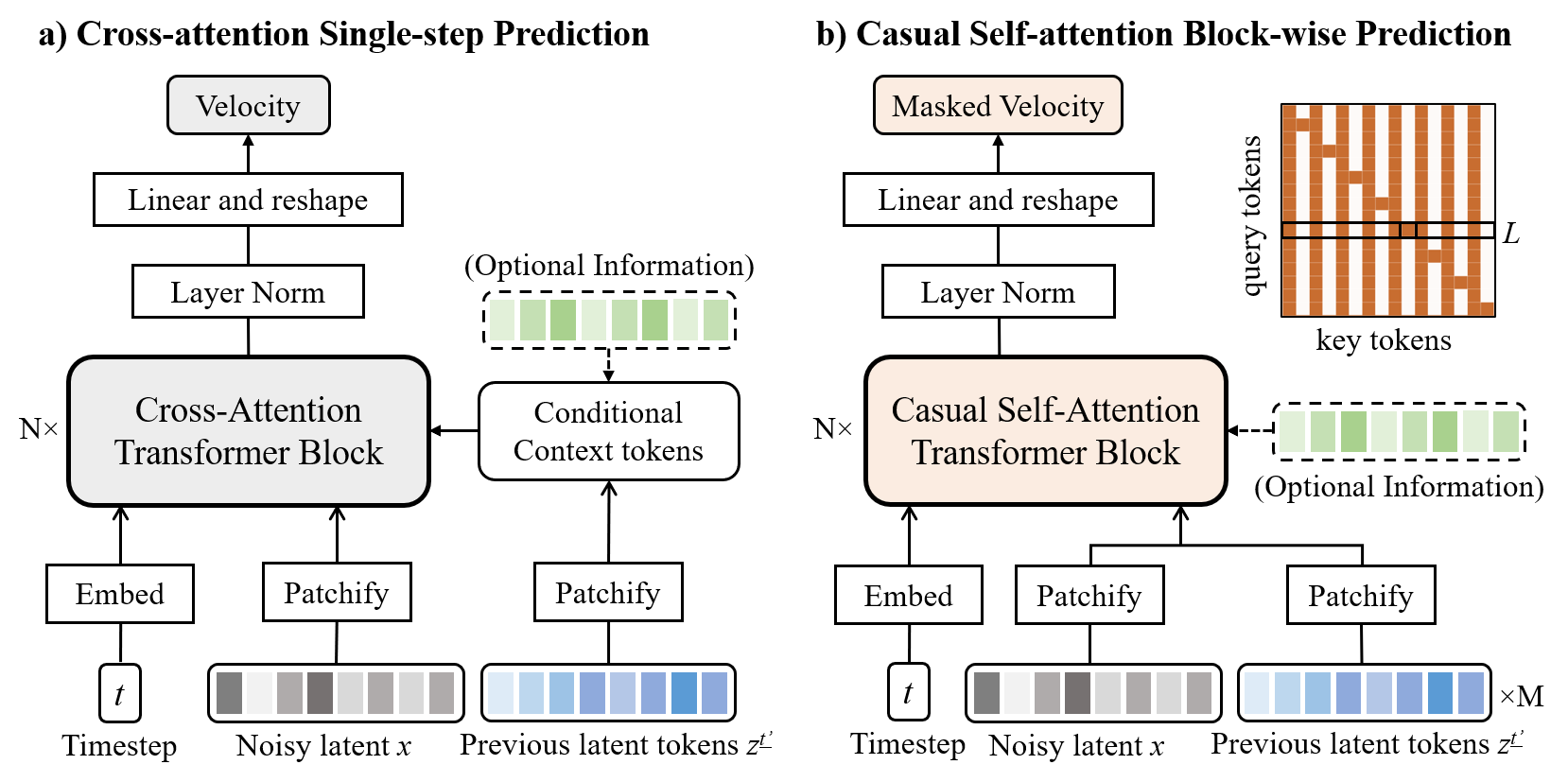}
    \caption{Illustration of latent-space autoregressive generative paradigms}
    \label{fig:latent_model}
\end{figure}

After mapping the original physical-space states into a compact latent token space, we apply flow matching to learn a transport from a Gaussian distribution to the latent data distribution, enabling autoregressive generation in the latent space. Specifically, we develop two prediction methods by modifying the scalable interpolant transformer \citep{ma2024sit} (SiT) backbone, as illustrated in Fig.~\ref{fig:latent_model}. In the first method, the previous two latent states are patchified into tokens and used as conditional context through the cross-attention mechanism of the transformer block for next-step prediction. In addition, auxiliary information, in particular the flow matching velocity field predicted by the preceding flow stage, is incorporated into the conditional context, providing the model with richer physics-relevant guidance. We refer to this method as GeoLAMP-S in Fig.~\ref{fig:latent_model}(a), where “S” denotes single-step prediction. In the second method, GeoLAMP-B, each conditioning frame and each noisy target frame is patchified into a spatial token group. We train with \(M=8\) conditioning frames and \(M=8\) target frames. The context-to-target mask lets conditioning groups attend to all conditioning groups; each target group attends to every conditioning group and its own spatial tokens, but not to other target groups (Fig.~\ref{fig:latent_model}(b)). Thus each flow-matching vector-field evaluation covers all \(M\) future latent frames in parallel. Autoregression occurs between completed blocks: committed predictions become context for the next block. 
The frame-wise embedders, transformer layers, and output head are reused across groups, so inference with different conditioning and output lengths changes the token stream and parameter-free mask without changing learned weights. Section~\ref{Inferencetimerolloutflexibility} measures the resulting accuracy tradeoff. Further flow-matching details are in Appendix~\ref{flowmatching}.

\subsection{Grid-based flow prior}
\label{sec:prior}
 We compute a cheap, low-resolution flow field on a regular grid and fuse it into the encoder, as illustrated in Fig.~\ref{geolamp_schm}(b). Given a binary structure at resolution \(G\), we solve the incompressible Navier--Stokes equations explicitly using \emph{fixed} convolution kernels rather than learned ones. Spatial derivatives of any field \(\psi\) on the grid are realized by three second-order stencils applied as convolutions,
\begin{equation}
\partial_{x}\psi \approx K_{x}\!*\!\psi,\qquad
\partial^{2}_{xx}\psi \approx K_{xx}\!*\!\psi,\qquad
\partial^{2}_{xy}\psi \approx K_{xy}\!*\!\psi,
\label{eq:stencils}
\end{equation}
with the central difference \(K_{x}=[-1,0,1]/2\Delta\), the Laplacian stencil \(K_{xx}=[1,-2,1]/\Delta^{2}\), and the four-corner mixed stencil \(K_{xy}=[\pm 1]/4\Delta^{2}\), and their transposes in \(y\). Applied to $\vec{u}$ and \(p\), these kernels yield all advective, viscous and pressure terms as batched convolutions, and the fields are marched explicitly to steady state on the coarse grid at negligible cost relative to the full simulation (details in Appendix~\ref{appen_c_method}). The converged field is compressed into a prior tensor \(\Phi\) carrying the two velocity components, the velocity magnitude and the pressure. Because averaging across a wall would invent flow through solid, \(\Phi\) is downsampled to the \(32\times32\) latent grid by a \emph{mask-aware} pooling that averages over fluid cells only,
\begin{equation}
\operatorname{pool}(\Phi)_{j}=\frac{\sum_{c\in\mathcal{C}_{j}}\Phi_{c}\,m_{c}}{\sum_{c\in\mathcal{C}_{j}}m_{c}},
\qquad
\phi_{j}=\frac{1}{|\mathcal{C}_{j}|}\sum_{c\in\mathcal{C}_{j}}m_{c},
\label{eq:maskpool}
\end{equation}
where \(\mathcal{C}_{j}\) are the fine cells pooled into latent cell \(j\) and \(m_{c}\in\{0,1\}\) is the fluid indicator. The pooled prior is concatenated onto the grid features produced by the GNO encoders and mixed by convolution.
\begin{equation}
\tilde{g}_{j}=W\big[\,g_{j}\,\Vert\,\operatorname{pool}(\Phi)_{j}\,\Vert\,\phi_{j}\,\big],
\label{eq:priorfuse}
\end{equation}
This grid--graph fusion supplies the global pressure and flow structure that the local geometric features alone cannot express. The fused encoder then predicts the steady velocity field from geometry alone, and that predicted field is what conditions the autoregressive transport rollout; this is the sense in which flow and transport are decoupled, and the two stages are evaluated separately in Section~\ref{sec:acc}.

\section{Experiment}
\subsection{Model accuracy}
\label{sec:acc}
\vspace{-5pt}
\paragraph{Geometry-to-velocity prediction} The steady velocity field is predicted first, directly from geometry, and only then conditions the autoregressive stage. We therefore evaluate this first stage on its own. We compare against the same five geometry-aware backbones, each trained \emph{prior-matched}, i.e.\ receiving exactly the coarse-prior and fluid-fraction channels that GeoLAMP's encoder consumes. Results for the two velocity components and the full velocity vector are reported in Table~\ref{tab:rel_l2_velocity}. GeoLAMP attains the lowest error on every component of both datasets. The coarse prior lowers the error of every model; the geometry-only ablation and example predictions are given in Appendix~\ref{appen_c_results}. Under both scenario, GeoLAMP exhibits the best performance on tortuous flow predictions.
\vspace{-10pt}
\begin{table}[H]
\centering
\small
\caption{Relative $L_2$ error for geometry-to-velocity prediction with the coarse flow prior. $u$ and $v$ are the streamwise and transverse components; $|\mathbf{u}|$ is the error of the full velocity vector, $\|\mathbf{u}_{\mathrm{pred}}-\mathbf{u}\|/\|\mathbf{u}\|$. Best/second-best are \textbf{bold}/\underline{underlined}.}
\label{tab:rel_l2_velocity}
\begin{tabular}{lccc|ccc}
\toprule
& \multicolumn{3}{c|}{Reactive Flow} & \multicolumn{3}{c}{Heat Convection} \\
\cmidrule(lr){2-4} \cmidrule(lr){5-7} 
Model & $u$ & $v$ & $|\mathbf{u}|$ & $u$ & $v$ & $|\mathbf{u}|$ \\
\midrule
OFormer \citep{li2023oformer}       & \underline{0.125} & \underline{0.183} & \underline{0.143} & 0.503 & 0.665 & 0.546 \\
GINO \citep{li2023gino}             & 0.206 & 0.278 & 0.229 & 0.450 & 0.569 & 0.482 \\
Transolver \citep{wu2024transolver} & 0.131 & 0.199 & 0.153 & \underline{0.364} & \underline{0.485} & \underline{0.396} \\
Geo-FNO \citep{li2023geofno}        & 0.142 & 0.213 & 0.165 & 0.382 & 0.495 & 0.412 \\
NUNO \citep{liu2023nuno}            & 0.451 & 0.573 & 0.488 & 0.568 & 0.690 & 0.601 \\
GeoLAMP                             & \textbf{0.097} & \textbf{0.135} & \textbf{0.109} & \textbf{0.177} & \textbf{0.246} & \textbf{0.196} \\
\bottomrule
\end{tabular}
\end{table}

\paragraph{Autoregressive prediction} Given the predicted velocity field, the transport dynamics are then rolled out in latent space. We conduct autoregressive experiments on three datasets using 4096 global--local (G\&L) points sampled from the method in Section.~\ref{gl}. We compare GeoLAMP with recent methods and evaluate performance using the relative $L_2$ error over the full autoregressive rollout as well as at the final prediction step, as reported in Table~\ref{tab:rel_l2_unified}. For both single-step and block-wise autoregression, GeoLAMP achieves the best last-step prediction accuracy across all three datasets, demonstrating its effectiveness in long-horizon forecasting. GeoLAMP-B's last-step errors exceed its rollout means but remain the lowest among the blockwise methods on all three datasets. Comparing GeoLAMP-S with the matched Latent-DiT control probes flow matching against DDPM noise prediction. Comparing GeoLAMP-B with Latent-Det. probes generative flow matching against direct latent-block regression; the largest last-step gap occurs on Elasticity.
\vspace{-7pt}
\begin{table}[H]
\centering
\small
\caption{Relative $L_2$ errors for autoregressive mean and last-step predictions on three datasets. Best/second-best results are \textbf{bold}/\underline{underlined}. The -S/-B labels denote the single-step/blockwise protocols evaluated here.}
\label{tab:rel_l2_unified}
\begin{tabular}{lcccccc}
\toprule
& \multicolumn{2}{c}{Heat Convection} 
& \multicolumn{2}{c}{Reactive Flow} 
& \multicolumn{2}{c}{Elasticity} \\
\cmidrule(lr){2-3} \cmidrule(lr){4-5} \cmidrule(lr){6-7}
Model 
& Mean & Last 
& Mean & Last 
& Mean & Last \\
\midrule

(-S) OFormer \citep{li2023oformer}
& 0.0060 & 0.0134 
& \underline{0.1101} & \underline{0.1384} 
& 9.29 & 144.4 \\

(-S) GINO \citep{li2023gino}
& 0.0050 & 0.0074 
& 0.1535 & 0.1675 
& 6.31 & 143.5 \\

(-S) Transolver \citep{wu2024transolver}
& 0.0052 & 0.0266
& 0.4766 & 0.4717
& \textbf{0.2520} & 1.3703 \\

(-S) Latent-DiT
& \underline{0.0037} & \underline{0.0073}
& 0.1464 & 0.1529
& 0.4340 & \underline{0.5892} \\

(-S) GeoLAMP
& \textbf{0.0033} & \textbf{0.0063}
& \textbf{0.1071} & \textbf{0.1183}
& \underline{0.4081} & \textbf{0.5154} \\

\midrule

(-B) Geo-FNO \citep{li2023geofno}
& 0.0041 & 0.0071
& 0.1369 & 0.1580
& \underline{0.3676} & \underline{0.5420} \\

(-B) NUNO \citep{liu2023nuno}
& 0.0043 & 0.0073
& 0.1528 & 0.1514
& 0.7160 & 3.8620 \\

(-B) Latent-Det.
& \underline{0.0035} & \underline{0.0056}
& \underline{0.0924} & \underline{0.1030}
& 0.4395 & 1.9324 \\

(-B) GeoLAMP
& \textbf{0.0025} & \textbf{0.0046} 
& \textbf{0.0789} & \textbf{0.0813} 
& \textbf{0.3460} & \textbf{0.3823} \\

\bottomrule
\end{tabular}
\end{table}
\vspace{-10pt}
\subsection{Effects of geometry representation}
\label{sec:georep}
\begin{figure}[H]
    \centering
    \includegraphics[width=0.95\linewidth]{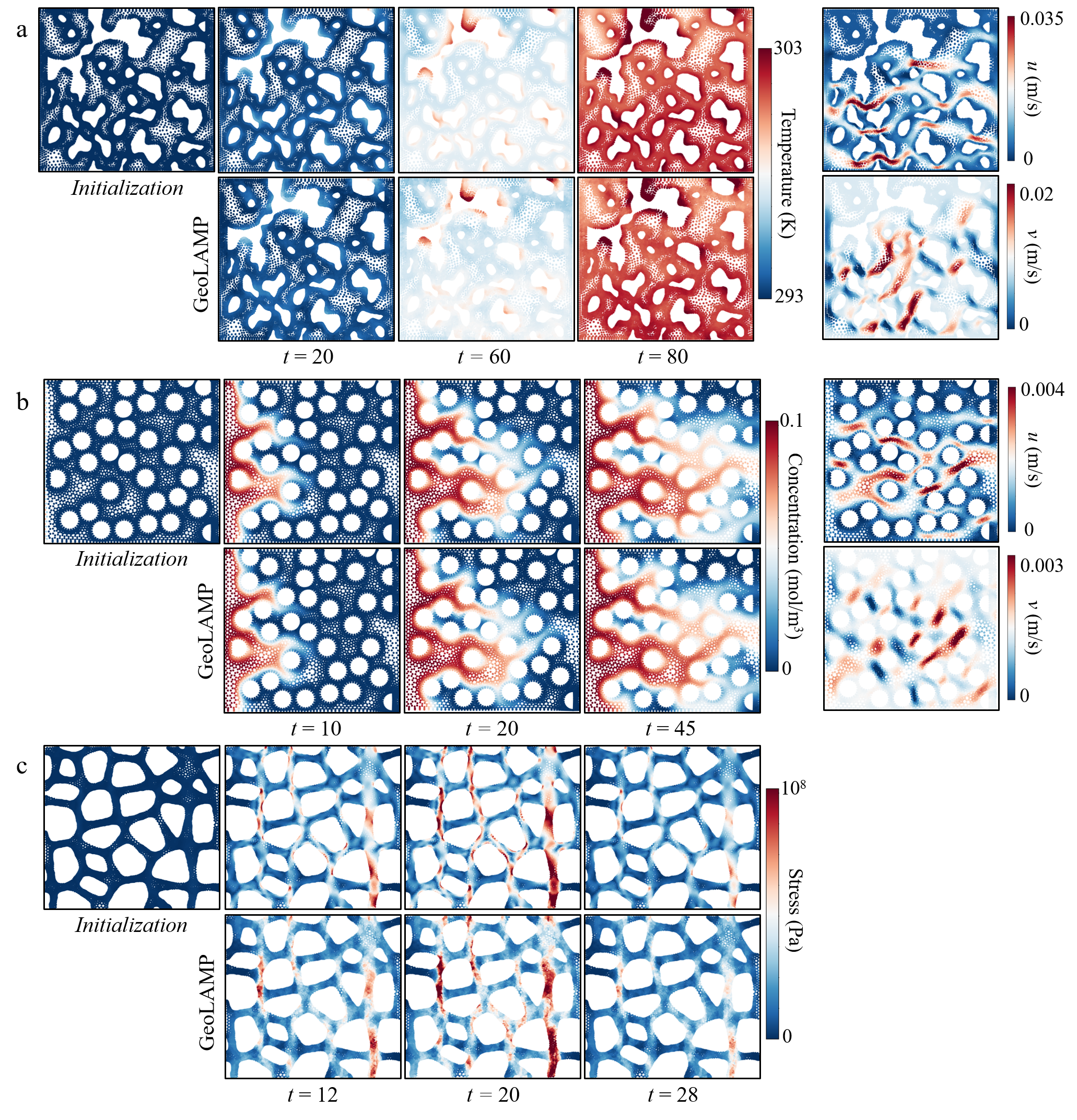}
    \vspace{-8pt}
    \caption{GeoLAMP results for velocity prediction (top for streamwise velocity prediction and bottom for transverse velocity prediction) and autoregression (bottom for prediction vs. top for ground truth). (a) Heat convection in stochastic-field structure. (b) Reactive flow in sphere-packed structure. (c) Elasticity in foam structure.}
    \label{fig:phys_rslt}
\end{figure}

We further examine the contribution of the proposed GE \& LE architecture to autoregressive prediction and evaluate the model's ability to reconstruct physical fields on full-mesh point sets.
\vspace{-5pt}
\paragraph{Designed global \& local encoder} To demonstrate the effectiveness of the proposed global \& local dual-encoder desgin, we compare GeoLAMP with a variant that encodes purely randomly sampled 4096 points (RS). The autoregressive accuracy on the three datasets is reported in Table~\ref{tab:rel_l2_sampledesign}. The higher accuracy on G\&L points demonstrates that carefully designed point selection is essential for bridging real and latent space, and is crucial for accurate latent-space prediction.
\vspace{-5pt}
\paragraph{Full-mesh generation by arbitrary decoder} Originating from the flexibility of the GNO-based decoder, the AD enables the generated latent states to be decoded back to real-space fields on arbitrary points. We further evaluate the performance using both the RS and the G\&L point sets. The autoregression results from G\&L points for three datasets are shown in Fig.~\ref{fig:phys_rslt}. The accuracy is given in Table~\ref{tab:rel_l2_sampledesign}, the G\&L point set not only achieves higher accuracy at its own locations, but also leads to better full-mesh recovery. The whole comparison of results with RS and G\&L points highlight a key distinction in scientific PDEs learning: spatial representation directly influence solution accuracy. This effect is further amplified in complex geometries, where physical behavior can vary sharply across subregions due to tortuous pathways and localized geometric features.
\vspace{-10pt}
\begin{table}[H]
\centering
\small
\caption{Relative $L_2$ error comparison across different sampling method and corresponding full-mesh recovery. We report the autoregression mean and the last-step error. Best results are in \textbf{bold}.}
\label{tab:rel_l2_sampledesign}
\begin{tabular}{lcccccc}
\toprule
& \multicolumn{2}{c}{Heat Convection} 
& \multicolumn{2}{c}{Reactive Flow} 
& \multicolumn{2}{c}{Elasticity} \\
\cmidrule(lr){2-3} \cmidrule(lr){4-5} \cmidrule(lr){6-7}
Locations 
& Mean & Last 
& Mean & Last 
& Mean & Last \\
\midrule

RS
& 0.0045 & 0.0088 
& 0.1178 & 0.1262 
& \textbf{0.3459} & \textbf{0.3811} \\

G\&L 
& \textbf{0.0025} & \textbf{0.0046} 
& \textbf{0.0789} & \textbf{0.0813} 
& 0.3460 & 0.3823 \\

\midrule

Full mesh from RS 
& 0.0045 & 0.0080 
& 0.1152 & 0.1349 
& 0.5351 & 0.5573 \\

Full mesh from G\&L 
& \textbf{0.0040} & \textbf{0.0077}
& \textbf{0.1065} & \textbf{0.1247}
& \textbf{0.3561} & \textbf{0.3684} \\

\bottomrule
\end{tabular}
\end{table}


\subsection{Inference-time rollout flexibility}
\label{Inferencetimerolloutflexibility}
GeoLAMP's causal block-wise attention allows the conditioning length $C$ and output length $P$ to be varied at inference time without retraining.
Because frame groups share model parameters, changing their number requires only rebuilding the parameter-free attention mask. We evaluate this flexibility using the frozen pretrained GeoLAMP-B models on 200 held-out trajectories per dataset, reporting errors normalized by the native $C=P=8$ configuration (Table~\ref{tab:block_flex}).

With $C=8$, increasing $P$ to $10$ or $12$ increases error by at most $1\%$ across datasets. Doubling $P$ to $16$ preserves near-native error for heat convection ($1.01\times$) and elasticity ($0.977\times$), but increases error to $1.42\times$ native for reactive flow. Thus, larger output blocks can reduce the number of autoregressive rollout iterations while maintaining near-native accuracy on some datasets.
With $P=8$, shortening the conditioning history to $C=6$ or $7$ increases error by at most $6\%$ or $2\%$, respectively. However, using a single conditioning frame increases error to $2.74\times$ native for heat convection and $2.04\times$ for reactive flow, compared with only $1.04\times$ for elasticity. These results reveal distinct sensitivities to output block length and conditioning history. Inference-time flexibility therefore enables task-dependent configuration choices, rather than uniform robustness to arbitrary block lengths.
\begin{table}[H]
\centering
\small
\caption{Inference-time block-length flexibility of pretrained GeoLAMP-B models. Values are horizon-averaged physical relative $L_2$ errors normalized by the corresponding native $C=P=8$ rollout (lower is better; $1.00$ is native). Left: output length at $C=8$, with commit stride $P$. Right: conditioning length at $P=8$.}
\label{tab:block_flex}
\begin{tabular}{lccccc|cccc}
\toprule
& \multicolumn{5}{c|}{Output length ($C=8$)}
& \multicolumn{4}{c}{History length ($P=8$)} \\
\cmidrule(lr){2-6} \cmidrule(lr){7-10}
Dataset
& $P{=}8$ & $P{=}10$ & $P{=}12$ & $P{=}14$ & $P{=}16$
& $C{=}1$ & $C{=}4$ & $C{=}6$ & $C{=}7$ \\
\midrule
Heat convection & 1.00 & 0.991 & 0.991 & 0.994 & 1.01  & 2.74 & 1.07 & 1.01 & 1.00 \\
Reactive flow   & 1.00 & 1.00  & 1.01  & 1.14  & 1.42  & 2.04 & 1.19 & 1.06 & 1.02 \\
Elasticity      & 1.00 & 0.964 & 0.993 & 0.986 & 0.977 & 1.04 & 1.07 & 1.03 & 1.01 \\
\bottomrule
\end{tabular}
\end{table}

\subsection{Attention variations along the temporal axis}
\label{subsec:attn}
\begin{figure}[H]
    \centering
    \includegraphics[width=1.0\linewidth]{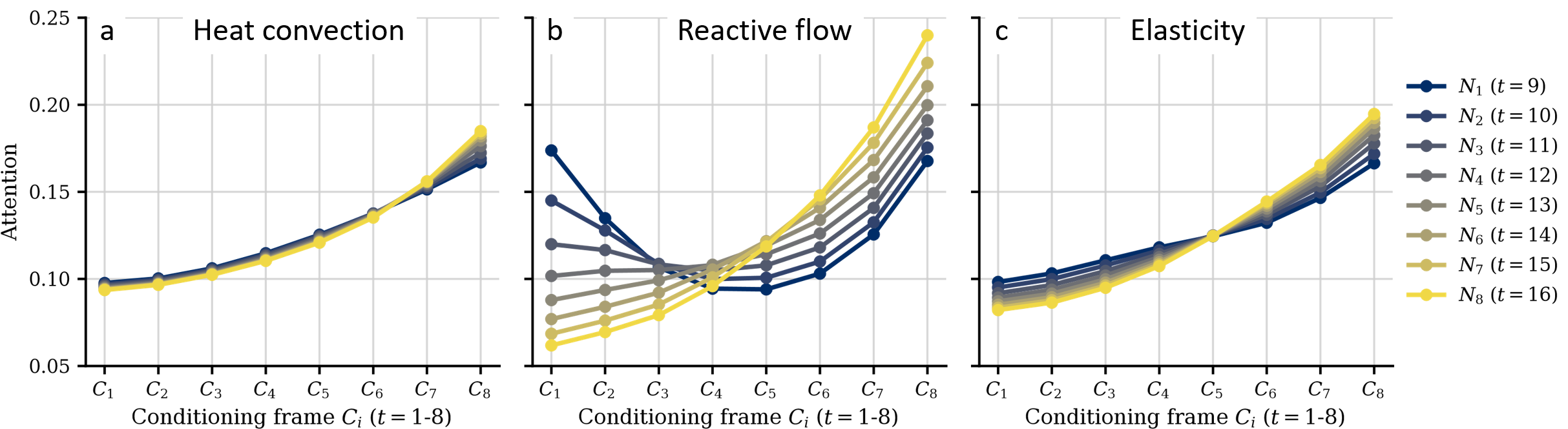}
    \vspace{-10pt}
    \caption{Attention strength along the temporal axis for the three dynamics.}
    \label{fig:attention_temporal}
\end{figure}

We analyze temporal changes by isolating the attention from noisy target queries $N_j$ to conditioning frames $C_i$. For each query, the attention weights renormalized over $C_1$--$C_8$ are shown as line plots in Fig.~\ref{fig:attention_temporal} (a)--(c), measuring the relative preference among conditioning frames. For heat convection and elasticity, temperature and stress increase continuously across the entire domain throughout the process, as shown in Fig.~\ref{fig:phys_rslt}. As a result, nearby states remain spatiotemporally consistent and continuous, gaining stronger attention weights. In contrast, reactive flow displays a different pattern. For near-future prediction, the initial conditioning frames remain nearly as important as the most recent frames. For later target frames, the attention gradually shifts toward recent inputs, while the contribution from the earliest frames decreases substantially. This behavior likely originates from the physics. As shown in Fig.~\ref{fig:phys_rslt}, the reaction fronts preferentially migrate along different paths of lower resistance, exhibiting time-lagged spatial heterogeneity. Together, these observations suggest that capturing long-term dependencies could be more beneficial for PDEs autoregressive learning.

\section{Conclusions}
\vspace{-4pt}
In this work, we propose GeoLAMP for multiphysics PDEs learning in meso-scale complex 2D geometries. The specially designed dual encoder enables the model to capture both global physical behavior and local features. By decoupling flow from transport, the velocity field is predicted from geometry before the transport dynamics are rolled out, and fusing a cheap grid-based flow prior into the graph encoder enhances tortuous flow predictions. We propose a casual self-attention transformer in a flow matching framework, and achieve block-wise autoregressive prediction. We construct three complex-geometry datasets, including heat convection, reactive flow, and elasticity. GeoLAMP achieves best accuracy in long-time autoregression and flexible rollout policy, while demonstrating strong flexibility in geometry representation for tortuous structures. Several directions remain for future work. First, although the dual-encoder architecture improves geometry-aware latent representation, the model still faces challenges in reconstructing high-quality physical fields in highly complex geometries. Second, while block-wise inference improves inference efficiency, the training cost increases with the block size. We plan to address this by introducing more efficient attention mechanisms. Moreover, meso-scale multiphysics encompasses a broad range of scientific and engineering problems in 3D, which remain to be further explored. Overall, GeoLAMP demonstrates promising potential as a fast surrogate model in solving multiphysics in tortuous complex domains.


\subsection*{Ethics statement}
This work studies surrogate modeling of multiphysics PDEs on synthetic meso-scale geometries. All three datasets are generated by the authors using finite element method following settings in Appendix~\ref{Appen_A}, which will be released upon acceptance. The authors declare no conflicts of interest.

\subsection*{Reproducibility statement}

We have made the following efforts to support reproducibility. The full generation procedure for
all three datasets, including geometries, governing equations, material properties, boundary
conditions is described in Appendix~\ref{Appen_A}. Complete
architectural hyperparameters for GeoLAMP and every baseline are given in Appendix~\ref{Appen_B}, and all training
configurations (learning rate, weight decay, scheduler, and epochs) are listed in
Table~\ref{tab:train}. All experiments were run on a single NVIDIA A100 GPU.

\bibliographystyle{unsrtnat}
\bibliography{references}


\appendix

\section{Dataset Generation}
\label{Appen_A}
\begin{figure}[H]
    \centering
    \includegraphics[width=0.65\linewidth]{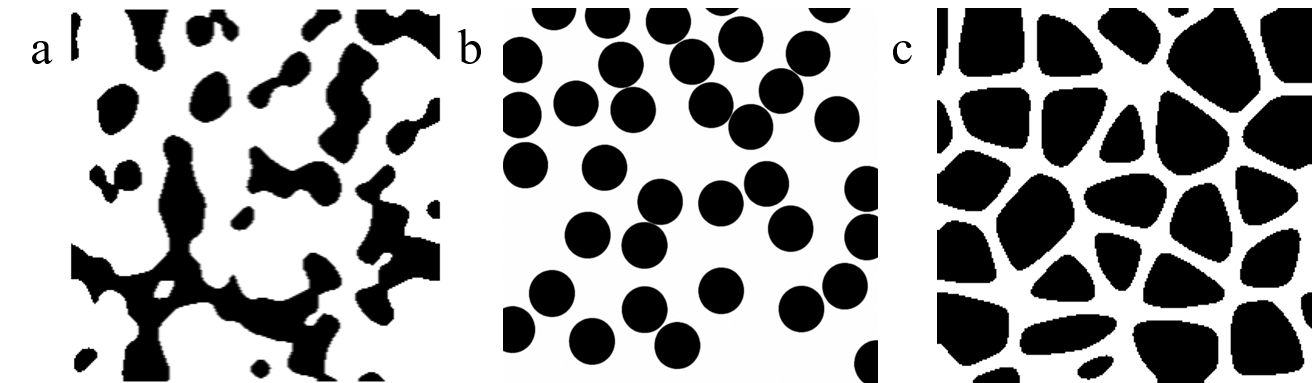}
    \caption{Meso-scale structure reconstruction}
    \label{fig:structure}
\end{figure}

\begin{figure}[H]
    \centering
    \includegraphics[width=0.95\linewidth]{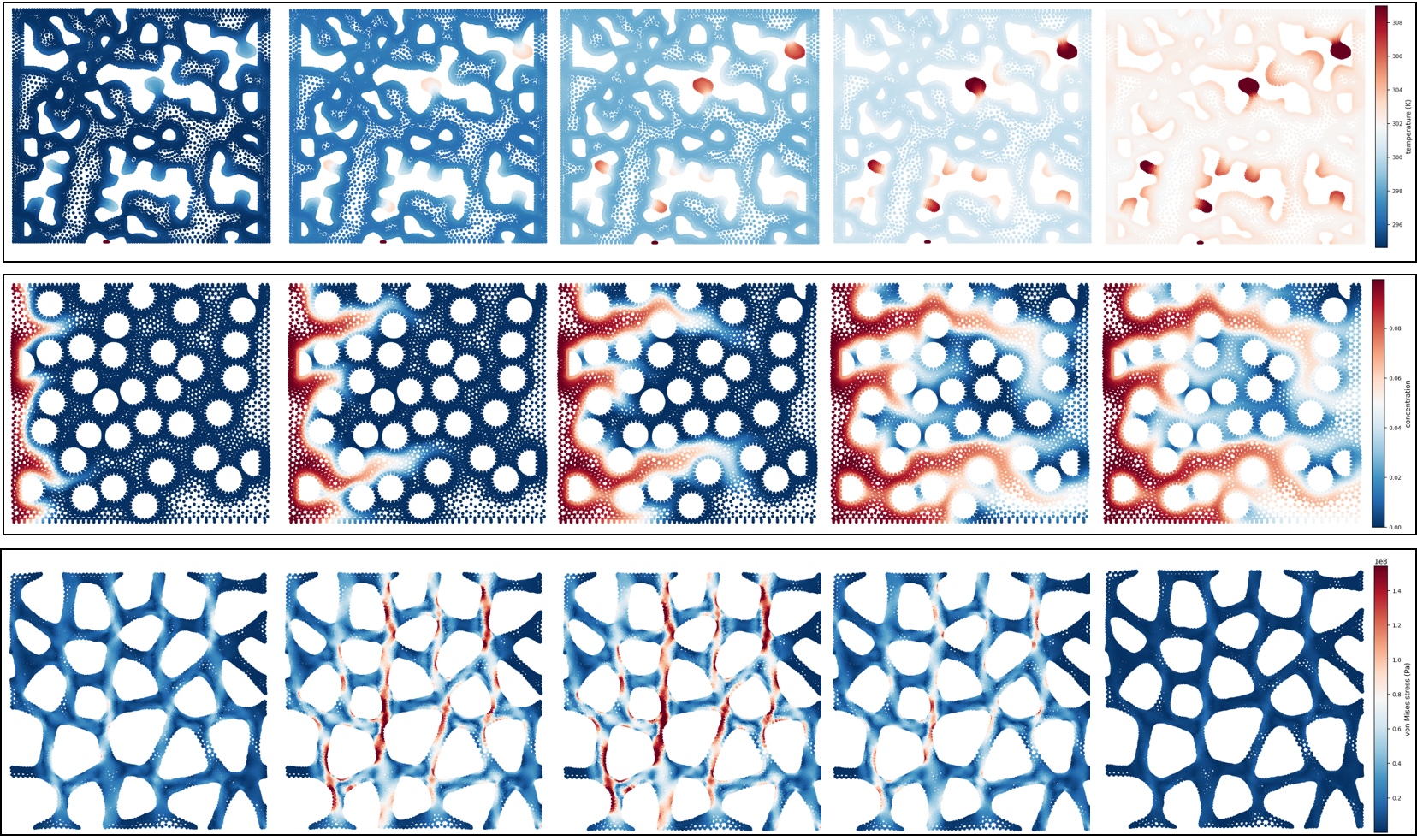}
    \caption{Meso-scale physics datasets}
    \label{fig:dataset_dist}
\end{figure}

Overall, we reconstruct three distinct meso-scale geometries for three different physical problems, as shown in Fig.~\ref{fig:structure}. For each dataset, 2,000 structures are generated, and the corresponding physical PDEs are solved to obtain 2,000 spatio-temporal solution sequences. For these different structures, the porosity changes from 0.55 to 0.70. Each dataset is then randomly split into training, validation, and test sets with 1,500, 300, and 200 sequences, respectively. All simulation is implemented in COMSOL \citep{multiphysics1998introduction}. We show one sequences of each dataset in Fig.~\ref{fig:dataset_dist}.
\paragraph{Heat convection in stochastic-field structure.}
The geometry is generated using a stochastic field method implemented in GeoDict \citep{ohser20093d}, as shown in Fig.~\ref{fig:structure}(a), yielding a highly heterogeneous porous structure within a $24{,}000\times24{,}000,\mu\mathrm{m}^2$ domain. The porosity of 2000 structures varies from 0.68 to 0.70. The flow is governed by the incompressible Navier--Stokes equations coupled with an energy transport equation, where heat exchange occurs at the solid--fluid interface. Water is used as the working fluid. Periodic boundary conditions are imposed along the flow direction with a pressure drop of $3,\mathrm{Pa}$, allowing the fluid to exit through the outlet and re-enter from the inlet while continuously exchanging heat with the solid phase under a prescribed heat flux of $1000,\mathrm{W}/\mathrm{m}^2$. In total, temperature field is saved at 100 time steps for the autoregression task. This setup represents pore-scale heat-transfer processes commonly encountered in battery systems and low-heat-flux thermal management applications.
\begin{equation}
\begin{array}{ll}
\rho\left(\frac{\partial \mathbf{u}}{\partial t} + \mathbf{u}\cdot\nabla \mathbf{u}\right)
= -\nabla p + \mu \nabla^2 \mathbf{u}, 
& x \in \Omega_f \\[3pt]

\nabla \cdot \mathbf{u} = 0, 
& x \in \Omega_f \\[3pt]

\frac{\partial T}{\partial t} + \mathbf{u}\cdot\nabla T 
= \alpha \nabla^2 T, 
& x \in \Omega_f \\[3pt]

- k_T \nabla T \cdot \mathbf{n} = q, 
& x \in \partial\Omega_s
\end{array}
\end{equation}

\paragraph{Reactive flow in sphere-packed structure.}
The complex geometry is constructed using a stochastic Monte Carlo sphere-packing process \citep{wang2022pore}, as shown in Fig.~\ref{fig:structure}(b). The porosity of 2000 structures varies from 0.61 to 0.68. Specifically, multiple spheres with radius $14,\mu\mathrm{m}$ are randomly initialized and iteratively adjusted within a $240\times240,\mu\mathrm{m}^2$ computational domain. In this domain, the incompressible Navier--Stokes equations are coupled with a solute transport equation, where reactants are advected from the left inlet to the right outlet and consumed at the solid surfaces through a surface reaction term. The inlet velocity, inlet concentration, and surface reaction rate are set to $7.2\times10^{-4},\mathrm{m}/\mathrm{s}$, $0.1,\mathrm{mol}/\mathrm{m}^3$, and $10^4,\mathrm{m}/\mathrm{s}$, respectively. In total, concentration field is saved at 60 time steps for the autoregression task. This setup captures key mechanisms in subsurface reactive transport and is representative of applications such as CO$_2$ sequestration and natural hydrogen generation.
\begin{equation}
\begin{array}{ll}
\rho\left(\frac{\partial \mathbf{u}}{\partial t} + \mathbf{u}\cdot\nabla \mathbf{u}\right)
= -\nabla p + \mu \nabla^2 \mathbf{u}, 
& x \in \Omega_f \\[3pt]

\nabla \cdot \mathbf{u} = 0, 
& x \in \Omega_f \\[3pt]

\frac{\partial c}{\partial t} + \mathbf{u}\cdot\nabla c 
= D \nabla^2 c, 
& x \in \Omega_f \\[3pt]

- D \nabla c \cdot \mathbf{n} = k c, 
& x \in \partial\Omega_s
\end{array}
\end{equation}
\paragraph{Elasticity in foam structure.}
The porous geometry is generated using a foam reconstruction method implemented in GeoDict \citep{ohser20093d}, as shown in Fig.~\ref{fig:structure}(c), within a $2{,}400\times2{,}400,\mu\mathrm{m}^2$ domain. The porosity of 2000 structures varies from 0.55 to 0.66. The mechanical response is modeled by infinitesimal linear elasticity under dynamic loading conditions. Copper is used as the solid material, reflecting its common use in metallic foams. Although the governing equations are quasi-static, time-dependent boundary displacements are applied to mimic realistic loading scenarios. In total, stress field is saved at 40 time steps. It should be mentioned that only first 38 timesteps are used, since the stress field at the last two steps approach zero, leading to analysis error. This dataset evaluates the model's ability to capture deformation responses under temporally varying boundary conditions in complex porous media.
\begin{equation}
\begin{array}{ll}
\nabla \cdot \sigma = 0, 
& x \in \Omega \\[3pt]

\sigma = \lambda (\nabla \cdot u) I + 2\mu \varepsilon, 
& x \in \Omega \\[3pt]

\varepsilon = \frac{1}{2}(\nabla u + (\nabla u)^{\top}), 
& x \in \Omega \\[3pt]

u(x,t) = -0.001 H \cdot \frac{1 - \cos(2\pi t)}{2}, 
& x \in \Gamma_{top}
\end{array}
\end{equation}

\section{Point selection examples}
\label{Appen_Pointselect}
\begin{figure}[h]
    \centering
    \includegraphics[width=\linewidth]{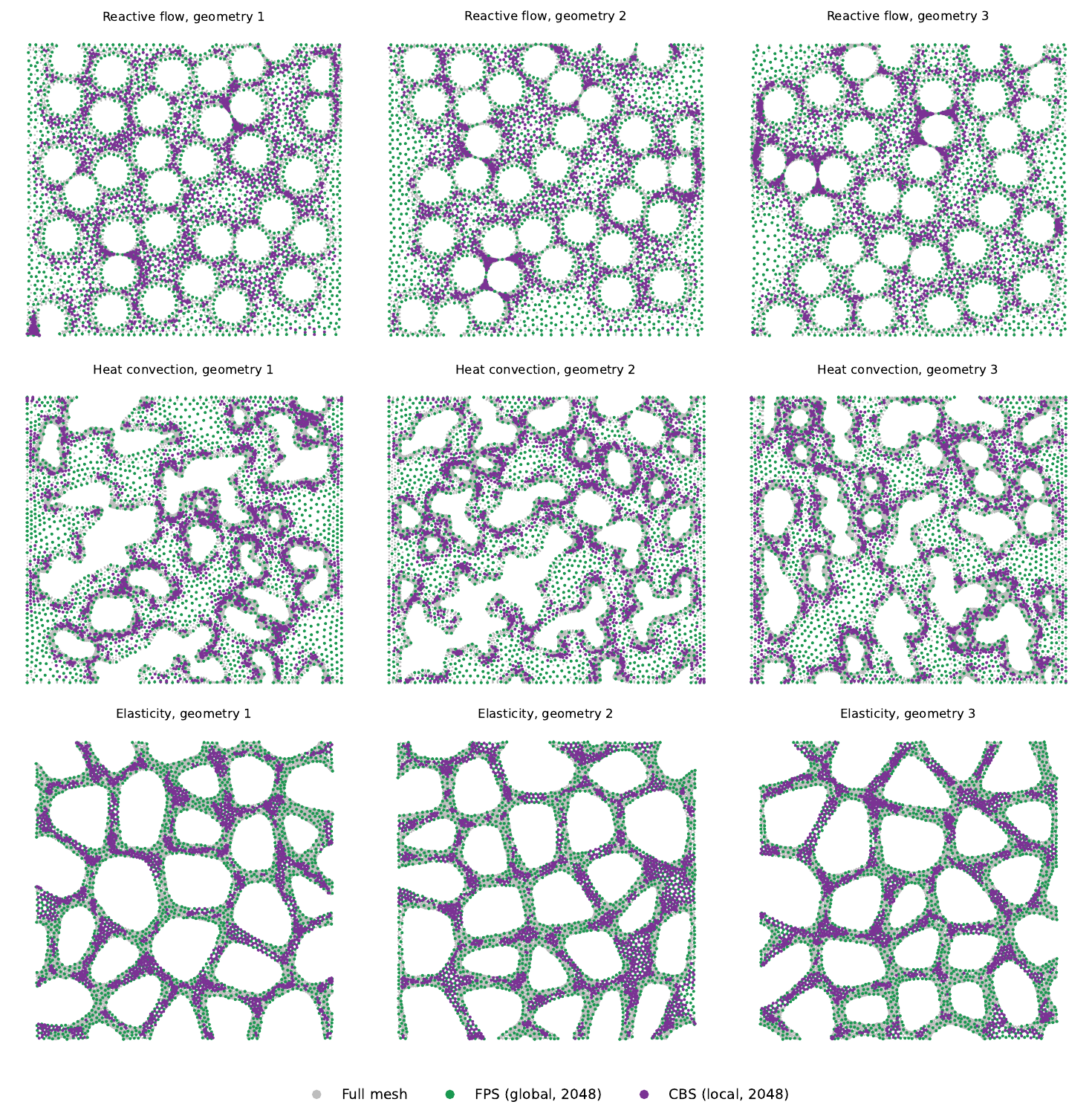}
    \caption{Global (FPS, green) and local (CBS, purple) point selections on the full mesh (gray) for three geometries of reactive flow (top), heat convection (middle) and elasticity (bottom). Each set contains 2048 points.}
    \label{fig:point_selection}
\end{figure}

Fig.~\ref{fig:point_selection} shows the global--local point sets of Section~\ref{gl} on three geometries from each dataset. FPS first draws $n_g=2048$ points from the full mesh, and CBS then selects $n_l=2048$ points with the largest scores from the remaining nodes, using $k=20$ nearest neighbors. The two sets are therefore disjoint. The point sets shown are the ones used in training. FPS spreads its points evenly over the whole domain, including the solid boundaries and the domain edges, and so provides the global backbone of the geometry. CBS points concentrate inside the local pores and tortuous channels, adds resolution where the global set is sparse rather than repeating it. The CBS points play an important role in learning the local physical details inside pores and channels, which carries critical information of tortuous flow-transport physics.

\section{Flow matching in GeoLAMP}
\label{flowmatching}
For both prediction paradigms, we train the latent generative model using a
  linear flow-matching objective. Let \(\mathbf{z}\) denote the clean target latent
  tokens and let \(\boldsymbol{\epsilon} \sim \mathcal{N}(\mathbf{0},\mathbf{I})\)
  denote standard Gaussian noise with the same shape as \(\mathbf{z}\). Given
  \(t \sim \mathcal{U}(0,1)\), we construct the interpolated latent state
  \begin{equation}
      \mathbf{x}_t = (1-t)\boldsymbol{\epsilon} + t\mathbf{z},
      \label{eq:linearinterpolation}
  \end{equation}
  which transports samples from the Gaussian distribution at \(t=0\) to the
  latent data distribution at \(t=1\). The corresponding target velocity is
  \begin{equation}
      \mathbf{u}_t = \frac{d\mathbf{x}_t}{dt}
      = \mathbf{z} - \boldsymbol{\epsilon}.
      \label{eq:targetvelocity}
  \end{equation}

  The neural network is trained to predict this velocity field from the noisy
  latent state, the flow-matching time, and the conditional context tokens
  \(\mathbf{c}\). The training objective is
  \begin{equation}
      \mathcal{L}_{\mathrm{FM}}
      =
      \mathbb{E}_{\mathbf{z},\boldsymbol{\epsilon},t}
      \left[
      \left\|
      \mathbf{v}_\theta(\mathbf{x}_t,t,\mathbf{c})
      -
      \mathbf{u}_t
      \right\|_2^2
      \right].
      \label{eq:lossfunction}
  \end{equation}

  For block-wise prediction, the clean latent variable \(\mathbf{z}\) corresponds
  only to the target block. Given a conditioning window
  \([\mathbf{z}_{s},\ldots,\mathbf{z}_{s+M-1}]\), the model predicts the target
  block
  \([\mathbf{z}_{s+M},\ldots,\mathbf{z}_{s+2M-1}]\). Flow matching is applied only
  to this target block, while the conditioning block is provided through
  \(\mathbf{c}\). Equivalently, the block-wise objective is an average over target
  latent positions:
  \begin{equation}
      \mathcal{L}_{\mathrm{FM}}^{\mathrm{block}}
      =
      \mathbb{E}_{\mathbf{z},\boldsymbol{\epsilon},t}
      \left[
      \frac{1}{|\Omega_{\mathrm{tar}}|}
      \sum_{i \in \Omega_{\mathrm{tar}}}
      \left\|
      \mathbf{v}_\theta(\mathbf{x}_t,t,\mathbf{c})_i
      -
      \mathbf{u}_{t,i}
      \right\|_2^2
      \right],
      \label{eq:blockloss}
  \end{equation}
  where \(\Omega_{\mathrm{tar}}\) denotes the set of target-block latent positions.
  In implementation, this target selection is performed implicitly by constructing
  \(\mathbf{x}_t\) only over the target block and by extracting velocity
  predictions only from the noisy target-token positions.

During inference, samples are generated by solving the learned probability-flow ODE
  \begin{equation}
      \frac{d\mathbf{x}_t}{dt}
      =
      \mathbf{v}_\theta(\mathbf{x}_t,t,\mathbf{c}),
      \qquad
      \mathbf{x}_0 \sim \mathcal{N}(\mathbf{0},\mathbf{I}),
      \qquad
      t \in [0,1].
      \label{eq:ode}
  \end{equation}
The terminal latent state \(\mathbf{x}_1\) gives the predicted target latent
block. For autoregressive rollout, the predicted latent states are appended to the conditioning window and the procedure is repeated. The terminal latents are then passed through the decoder to reconstruct the corresponding physical-space states. For the reported test rollouts, we use Euler integration with 50 discretization steps.

\section{Coarse flow prior and velocity prediction}
\label{Appen_C}
\subsection{Coarse flow prior: method and settings}
\label{appen_c_method}
The prior solver applies the fixed stencils of Eq.~\ref{eq:stencils} to \(u\), \(v\) and \(p\), so all advective, viscous and pressure terms are evaluated as batched convolutions on GPU. The velocity is marched explicitly to steady state; the pressure is updated by a five-point Jacobi iteration of \(\nabla^{2}p=-\rho(u_{x}^{2}+2u_{y}v_{x}+v_{y}^{2})\) with an added \(\nabla\!\cdot\!\mathbf{u}/\Delta t\) damping term. No-slip is imposed by masking solid cells, and a fixed pressure drop drives the flow between inlet and outlet buffers. The result is a coarse, explicit approximation of the physics solved in the dataset.

The grid-based flow prior of Section~\ref{sec:prior} is precomputed once per geometry and cached, so it adds no cost during training or inference. The solver settings used for all reported results are listed in Table~\ref{tab:prior}. The occupancy mask is rasterized at $G=256$ and the converged field is pooled to the $32\times32$ latent grid with an exact kernel of size $k=G/32=8$. Because the occupancy grid uses a square bounding box while the physical domain is not square, the padding rows above the domain are forced solid; leaving them fluid would give the solver a frictionless bypass channel along the top boundary and corrupt the pressure field.

Convergence is monitored by two criteria evaluated together: the relative change of the velocity field between checks, and a flux-balance measure computed as the interquartile spread of the column-wise flux over interior columns, normalized by its median. The columns adjacent to the inlet and outlet buffers are excluded from this measure, since the buffers are forced fully fluid and therefore carry artificially more flux. A sequence is accepted only if both criteria are met; sequences that fail, or that produce any non-finite value, are marked invalid and assigned a zero prior together with a zero fluid-fraction channel, so the encoder can distinguish a missing prior from a genuinely stagnant region. For the reported datasets, $99.6\%$ of sequences converged with a median flux imbalance of $0.011$, against an acceptance threshold of $95\%$ and $0.05$ respectively.

Two properties of the scheme are worth stating plainly. First, no-slip is imposed by masking solid cells rather than by a boundary discretization, so accuracy degrades near walls; the field is used as a carrier of global connectivity and pressure structure, and should not be read as a quantitative near-wall velocity. Second, the collocated grid admits a checkerboard pressure mode, which is monitored explicitly and was found to remain small at the reported settings.

\begin{table}[h]
\centering
\caption{Solver and pooling settings for the coarse flow prior.}
\label{tab:prior}
\small
\begin{tabular}{l c l}
\toprule
Setting & Value & Note \\
\midrule
Solve resolution $G$          & 256      & pooled to $32\times32$, kernel $k=8$ \\
Time step $\Delta t$          & 0.2      & explicit marching to steady state \\
Max iterations                & 100{,}000 & median convergence $\approx$ 5{,}000 \\
Velocity residual tolerance   & $10^{-4}$ & relative change between checks \\
Flux-balance tolerance        & 0.05     & interquartile spread / median \\
Inlet--outlet buffer          & 8 cells  & fixed pressure drop, fully fluid \\
Transverse boundary           & free-slip & preserves the $y$-flip symmetry \\
Validity gate                 & $\geq 95\%$ & achieved $99.6\%$ \\
\bottomrule
\end{tabular}
\end{table}

\subsection{Velocity prediction results}
\label{appen_c_results}
Table~\ref{tab:vel_noprior} reports the geometry-only counterpart of Table~\ref{tab:rel_l2_velocity}: every model is trained from geometry alone, without the coarse-prior channels. GeoLAMP is again the most accurate on every component, and its margin is largest in this setting (reactive flow $0.153$ vs.\ $0.263$ for Geo-FNO; heat convection $0.482$ vs.\ $0.608$). Without the prior, however, several baselines approach the $0.891$ error of a zero prediction and recover little beyond the mean field; on heat convection OFormer and Transolver reach $v\approx1.000$, i.e.\ they predict an almost vanishing transverse velocity. Adding the prior lowers the vector error of every baseline, by $1.4$--$4.2\times$ on reactive flow and $1.3$--$2.2\times$ on heat convection, and of GeoLAMP by $1.4\times$ and $2.5\times$ respectively.

\begin{figure}[H]
    \centering
    \includegraphics[width=0.7\linewidth]{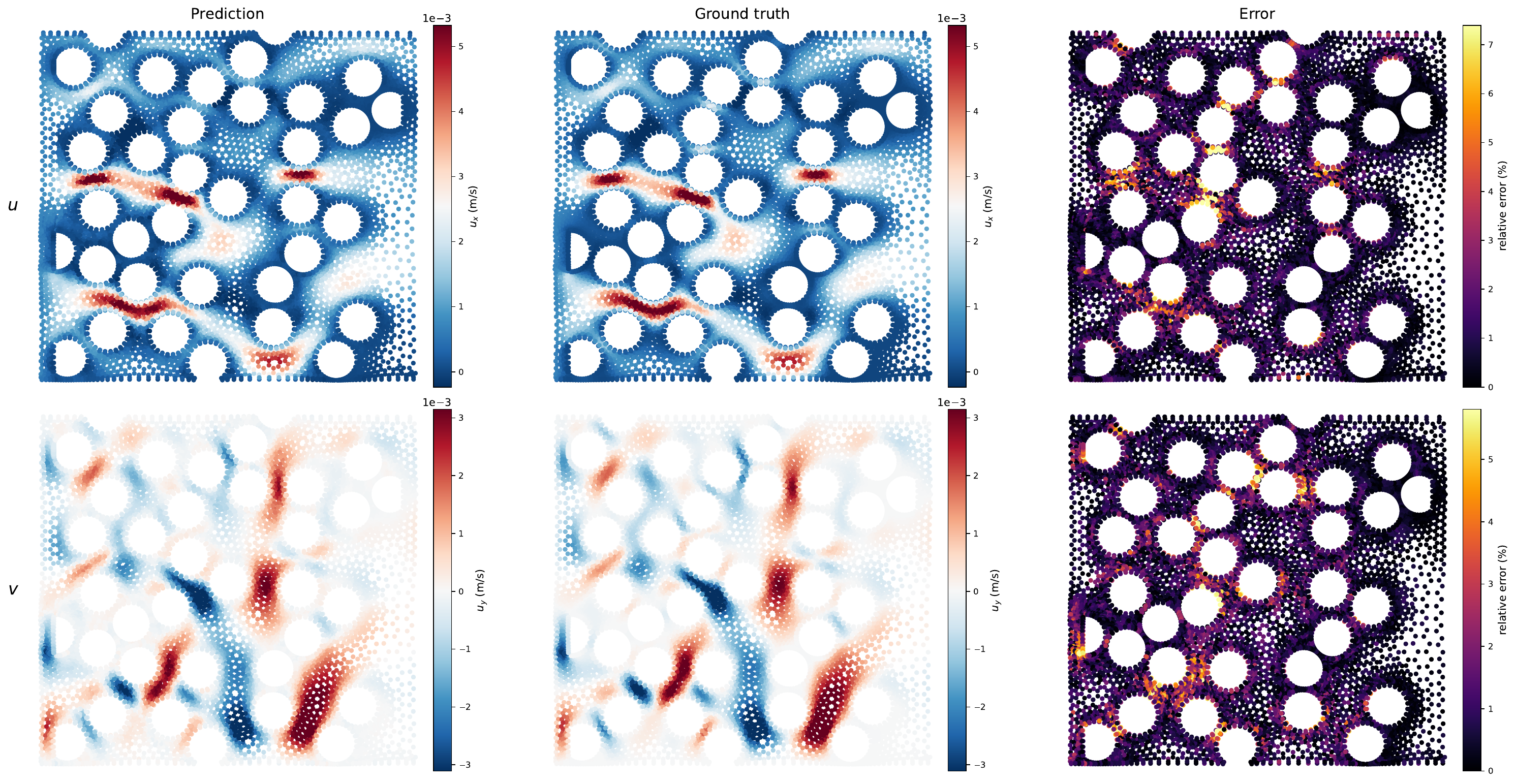}\\[-2pt]
    {\small (a) Reactive flow, test structure 400}\\[6pt]
    \includegraphics[width=0.7\linewidth]{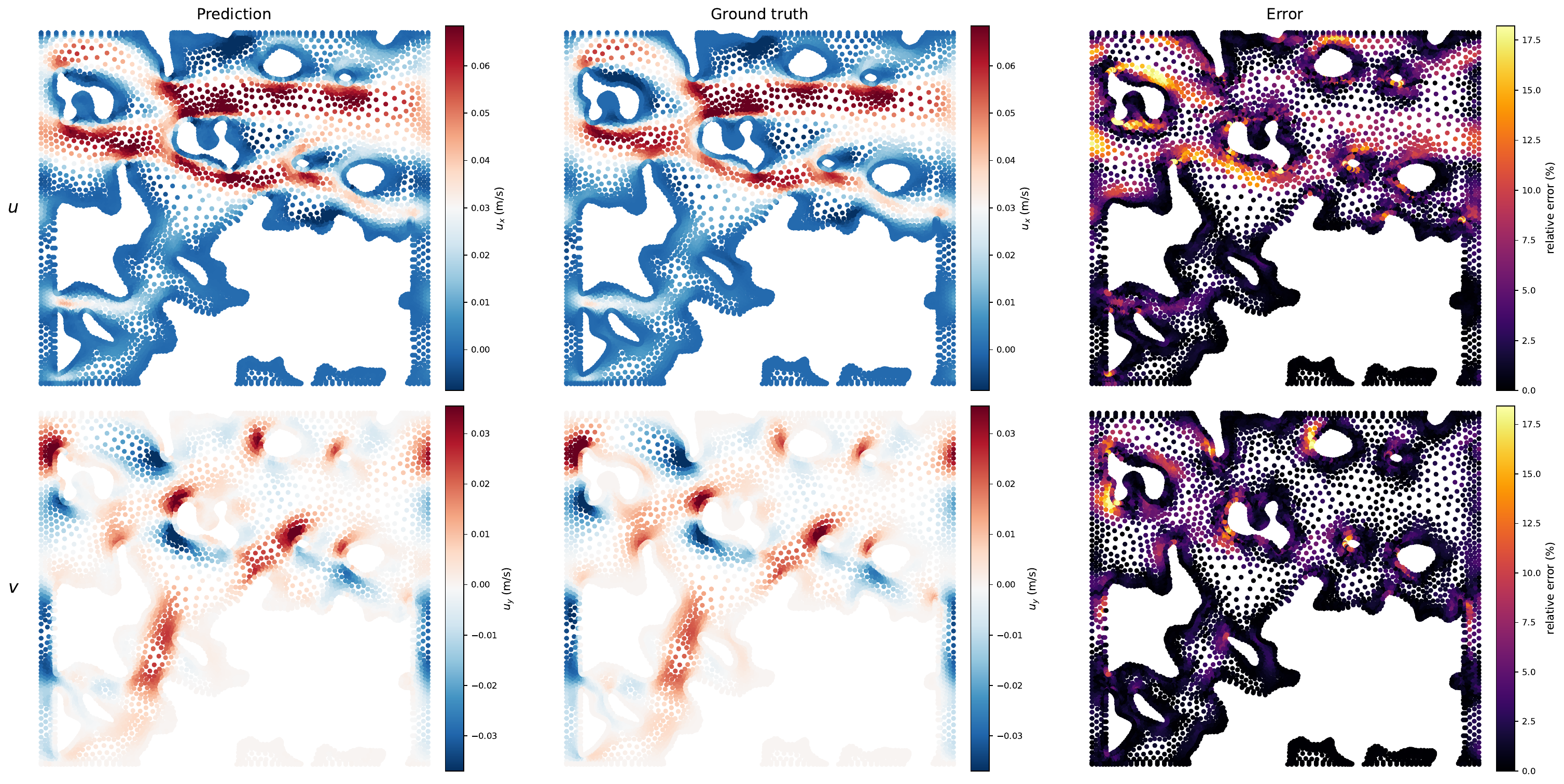}\\[-2pt]
    {\small (b) Heat convection, test structure 751}
    \caption{Geometry-to-velocity prediction of GeoLAMP with the coarse flow prior on held-out structures. In each panel, columns show the prediction, the ground truth and the pointwise relative error; rows show the streamwise ($u$) and transverse ($v$) velocity components.}
    \label{fig:vel_pred}
\end{figure}

Fig.~\ref{fig:vel_pred} shows GeoLAMP predictions with the prior on a held-out structure of each dataset. The preferential flow channels are reproduced in both components, and the largest errors concentrate at pore throats, where the masked no-slip treatment of the coarse prior is least accurate.

\begin{table}[h]
\centering
\small
\caption{Relative $L_2$ error for geometry-to-velocity prediction \emph{without} the coarse flow prior (geometry only). Columns as in Table~\ref{tab:rel_l2_velocity}. Best/second-best among geometry-only models are \textbf{bold}/\underline{underlined}.}
\label{tab:vel_noprior}
\begin{tabular}{lccc|ccc}
\toprule
& \multicolumn{3}{c|}{Reactive Flow} & \multicolumn{3}{c}{Heat Convection} \\
\cmidrule(lr){2-4} \cmidrule(lr){5-7} 
Model & $u$ & $v$ & $|\mathbf{u}|$ & $u$ & $v$ & $|\mathbf{u}|$ \\
\midrule
OFormer \citep{li2023oformer}       & 0.464 & 0.716 & 0.546 & 0.700 & 1.000 & 0.788 \\
GINO \citep{li2023gino}             & 0.415 & 0.583 & 0.468 & 0.722 & 0.879 & 0.764 \\
Transolver \citep{wu2024transolver} & 0.564 & 0.831 & 0.649 & 0.792 & 1.000 & 0.853 \\
Geo-FNO \citep{li2023geofno}        & \underline{0.229} & \underline{0.334} & \underline{0.263} & \underline{0.571} & \underline{0.710} & \underline{0.608} \\
NUNO \citep{liu2023nuno}            & 0.640 & 0.846 & 0.704 & 0.754 & 0.921 & 0.801 \\
GeoLAMP                             & \textbf{0.137} & \textbf{0.188} & \textbf{0.153} & \textbf{0.456} & \textbf{0.564} & \textbf{0.482} \\

\bottomrule
\end{tabular}
\end{table}

\section{Commit step influence}
\label{Append_Commit}
\begin{figure}[H]
    \centering
    \includegraphics[width=1\linewidth]{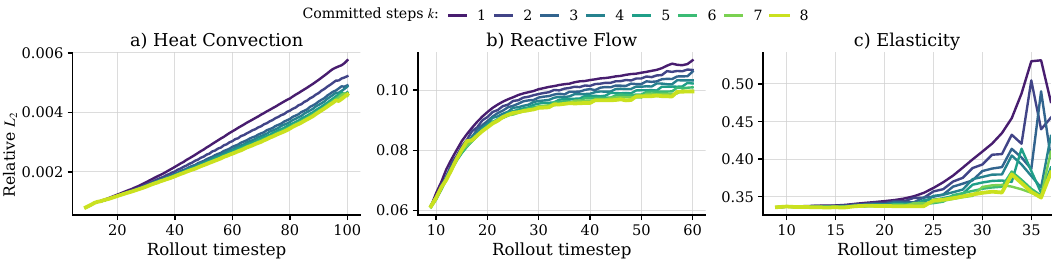}
    \caption{Relative $L_2$ rollout error for different committed steps during blockwise prediction.}
    \label{fig:8tok}
\end{figure}

The blockwise prediction paradigm provides the flexibility to commit different numbers of predicted steps during inference. Specifically, after predicting a block of future latent states, the model can commit only the first few steps and then re-predict the next block, or commit more steps at once before the next autoregressive update. We investigate this effect on three datasets by varying the number of committed steps $k$, as shown in Fig.~\ref{fig:8tok}.
GeoLAMP-B permits the commit stride $k$ to vary while keeping the predicted block length fixed. Across the three datasets, larger $k$ generally yields lower rollout error, particularly at later timesteps and for Elasticity. This trend is consistent with the reduced number of autoregressive updates when more predicted steps are committed at once.

\section{Model hyperparameter and implementation details}
\label{Appen_B}
\subsection{Model hyperparameter}
\paragraph{Global \& Local VAE.}
The autoencoder compresses each per-timestep scalar field on the unstructured mesh into a $64\times8\times8$ latent. Two independent GNO encoders use radius-graph integral transforms with radius $0.15$ and MLP layers $[80,80,80]$, projecting the input points onto a shared $32\times32$ grid with 128 channels. The two encoded grids are summed element-wise. The CNN encoder uses a base hidden width of 128, channel multipliers $(1,2,4)$, two ResNet blocks per resolution level, and self-attention at resolutions $16\times16$ and $8\times8$, producing an $8\times8\times64$ bottleneck with mean and log-variance. The KL weight is set to $10^{-6}$. The decoder mirrors the CNN encoder and reconstructs a $32\times32\times128$ grid, followed by a GNO decoder with MLP layers $[512,256]$ and radius $0.15$.

\paragraph{Arbitrary Decoder.}
The arbitrary decoder shares the CNN-decoder and GNO-decoder backend with the VAE. It takes a predicted $8\times8\times64$ latent and upsamples it through three resolution levels with base hidden width 128, channel multipliers $(1,2,4)$, two ResNet blocks per level, and self-attention at resolutions $8\times8$ and $16\times16$. The CNN decoder outputs a $32\times32\times128$ feature grid. A GNO decoder with MLP layers $[512,256]$ and radius $0.15$ maps the grid features to the full mesh with approximately $12$--$21\times10^3$ valid vertices and produces a one-channel field prediction.

\paragraph{GeoLAMP-S.}
GeoLAMP-S denoises a $64\times8\times8$ target latent under conditioning. With $\mathtt{patch\_size}=1$, each latent is patch-embedded into 64 tokens with hidden width 768 and sinusoidal 2-D positional embeddings. The flow-matching time and autoregressive step index are embedded using sinusoidal-MLP timestep embedders and combined as the AdaLN-Zero conditioning vector. The transformer uses 12 blocks, hidden width 768, 12 attention heads, and $\mathrm{mlp\_ratio}=4$. Each block contains AdaLN-modulated self-attention, cross-attention to context tokens, and an AdaLN-modulated GeGLU feed-forward layer. The final AdaLN-modulated linear layer maps the tokens back to 64-channel patches.

\paragraph{GeoLAMP-B.}
GeoLAMP-B uses the same backbone and hyperparameter settings as GeoLAMP-S, but performs non-overlapping block-wise prediction. It predicts $M=8$ future latent states in one flow-matching evaluation. Conditional and noisy target latents are patch-embedded into a causal token sequence, and the noisy target token groups are unpatchified and concatenated to produce the eight predicted latent states. The velocity here is the flow-matching vector field, not the physical velocity of Section~\ref{sec:acc}.

\paragraph{GINO.}
GINO uses the same encoder-decoder structure as the Global \& Local VAE (changed to AE), with the variational bottleneck replaced by a deterministic autoencoder, and are trained together. The FNO operates on the $8\times8$ latent grid. A $1\times1$ lifting convolution maps the input to a 752-channel hidden representation, using 192 input channels for reaction flow and heat convection and 128 input channels for elasticity. The trunk contains 10 spectral blocks with $(4,4)$ Fourier modes, complex-valued mode-wise weights, $1\times1$ pointwise convolutional residual bypasses, GELU activations, and outer residual connections. A final $1\times1$ projection convolution maps the hidden tensor to 64 output channels.

\paragraph{OFormer.}
OFormer operates directly on 4096 mesh nodes sampled by FPS and CBS. The encoder uses a $\mathrm{Conv2d}(5\to128)$ stem with kernel size $(2,1)$, learned node-type embeddings, and relative positional encodings. The encoder contains six Galerkin linear-attention blocks with hidden dimension 128, head dimension 128, one attention head, and feed-forward hidden dimension 512. The decoder uses Gaussian Fourier coordinate features, a two-layer MLP with hidden dimension 128, one Galerkin cross-attention block, one linear-attention block, and a three-layer MLP prediction head.

\paragraph{Transolver.}
Transolver operates directly on 4096 real-space query points from the concatenated FPS and CBS point clouds. The input feature dimension is 6 for reaction flow and heat convection and 4 for elasticity. A preprocessing MLP maps the input to width 256, followed by a learned placeholder vector and a sinusoidal physical-time embedding processed by a two-layer SiLU MLP. The trunk contains 6 Transolver blocks with hidden size 256, 8 heads, 32 slice tokens per head, dropout 0, and $\mathrm{mlp\_ratio}=2$. A final LayerNorm and linear head predict one scalar field value per point.

\paragraph{Geo-FNO.}
Geo-FNO learns a deformation from the irregular input domain to a regular computational grid and applies an FNO trunk on that grid, mapping back to the physical points through the inverse deformation. It is trained on all three datasets under the same blockwise $M$-to-$M$ ($M=8$) autoregression protocol, with matched model size, identical training strategy, and identical inference protocol as GeoLAMP-B, and is evaluated on the same 4096 G\&L points of the test set.

\paragraph{NUNO.}
NUNO partitions the non-uniform point set into subdomains, interpolates each subdomain onto a local regular grid, and applies an FNO per subdomain before scattering the predictions back to the original points. It follows the same blockwise $M$-to-$M$ ($M=8$) protocol, matched model size, training strategy, inference protocol, and 4096 G\&L evaluation points as Geo-FNO and GeoLAMP-B.

\paragraph{Latent-Det.}
Latent-Det.\ is a matched deterministic control for the generative formulation. It uses the same frozen Global \& Local VAE, the same latent tokens, the same causal block mask, the same $M=8$ blockwise protocol and commit stride, matched transformer depth and width, and identical conditioning as GeoLAMP-B. The flow-matching objective and its iterative ODE sampler are replaced by direct MSE regression of the same target latent block, so inference uses a single forward pass per block rather than 50 Euler steps.

\paragraph{Latent-DiT.}
Latent-DiT denoises a $64\times8\times8$ target latent using DDPM noise prediction. With $\mathtt{patch\_size}=1$, the noisy target latent is embedded into 64 tokens with hidden width 768 and fixed sinusoidal 2-D positional embeddings. The diffusion timestep is embedded using a sinusoidal-MLP timestep embedder for AdaLN-Zero modulation. The transformer uses 12 blocks, hidden width 768, 12 attention heads, and $\mathrm{mlp\_ratio}=4$. Each block contains AdaLN-modulated self-attention, cross-attention to context tokens, and an AdaLN-modulated GeGLU feed-forward layer. The final AdaLN-modulated linear layer maps the tokens back to 64-channel patches.

\subsection{Training settings}
The training configurations are summarized in Table~\ref{tab:train}. For the heat convection and reactive flow datasets, the input includes the velocity field as an optional context. For the elasticity dataset, the model takes only the von Mises stress field as input. The VAE and latent dynamics model are trained separately. The arbitrary decoder is also trained independently, using precomputed latent vectors from partial points as input and full-mesh data as the target. All training tasks are conducted on a single NVIDIA A100 GPU with 80 GB of memory.
\begin{table}[h]
\centering
\caption{Training hyperparameters for baseline models, GeoLAMP variants, the VAE, and the arbitrary decoder.}
\label{tab:train}
\small
\begin{tabular}{l c c l l c c}
\toprule
Model & Learning rate & Weight decay & Scheduler & Epochs \\
\midrule
GINO
  & $10^{-4}$ & $10^{-4}$
  & cosine, 500-step warmup
  & 200 \\
OFormer
  & $10^{-4}$ & $10^{-4}$
  & cosine, 500-step warmup
  & 200 \\
Transolver
    & $3{\times}10^{-4}$ & $10^{-4}$
    & cosine, 500-step warmup
    & 200 \\
GeoFNO
  & $10^{-4}$ & $10^{-4}$
  & cosine, 500-step warmup
  & 200 \\
NUNO
  & $10^{-4}$ & $10^{-4}$
  & cosine, 500-step warmup
  & 200 \\
Latent-DiT
    & $10^{-4}$ & $10^{-4}$
    & cosine, 500-step warmup
    & 200 \\
Latent-Det.
  & $10^{-4}$ & $10^{-4}$
  & cosine, 500-step warmup
  & 200 \\
GeoLAMP-S
  & $10^{-4}$ & $10^{-4}$
  & cosine, 500-step warmup
  & 200\\
GeoLAMP-B
& $2.5{\times}10^{-5}$ & $10^{-4}$
    & cosine, 1000-step warmup
    & 200 \\
VAE
  & $3{\times}10^{-4}$ & $10^{-4}$
  & ReduceLROnPlateau (factor 0.5, pat. 5)
  & 50 \\
AD
  & $10^{-4}$ & $10^{-4}$
  & ReduceLROnPlateau (factor 0.5, pat. 10)
  & 100\\
\bottomrule
\end{tabular}
\end{table}

\newpage

\end{document}